\documentclass[runningheads]{llncs}
\usepackage[T1]{fontenc}
\usepackage{graphicx,verbatim}
\usepackage{ makecell, graphicx}
\usepackage{amsmath}
\usepackage{multirow}
\usepackage{booktabs}
\usepackage[table]{xcolor} 
\usepackage{colortbl}      
\usepackage{hyperref}
\usepackage{amssymb}
\definecolor{baseblue}{RGB}{225,240,255}
\colorlet{colshade}{baseblue!30!white} 
\newcommand{\best}[1]{\textbf{#1}}   
\newcommand{\second}[1]{\underline{#1}}   

\newcommand\modelname{\textsf{RLAR}}

\begin{document}
\title{Representation-Level Adversarial Regularization for Clinically Aligned Multitask Thyroid Ultrasound Assessment}
\titlerunning{\modelname{} for Clinically Aligned Multitask Thyroid Ultrasound Assessment}
%

\author{
Dina Salama\inst{1} \and 
Mohamed Mahmoud\inst{2} \and 
Nourhan Bayasi\inst{1} \and 
David Liu\inst{1} \and 
Ilker Hacihaliloglu\inst{1}
}

\authorrunning{Dina et al.}

\institute{
University of British Columbia, Vancouver, Canada \and
University of Calgary, Calgary, Canada \\
\email{dinakhalid404@gmail.com}
}


  
\maketitle              
\begin{abstract}
Thyroid ultrasound is the first-line exam for assessing thyroid nodules and determining whether biopsy is warranted. In routine reporting, radiologists produce two coupled outputs: a nodule contour for measurement and a TI-RADS risk category based on sonographic criteria. Yet both contouring style and risk grading vary across readers, creating inconsistent supervision that can degrade standard learning pipelines. In this paper, we address this workflow with a clinically guided multitask framework that jointly predicts the nodule mask and TI-RADS category within a single model. To ground risk prediction in clinically meaningful evidence, we guide the classification embedding using a compact TI-RADS-aligned radiomics target during training, while preserving complementary deep features for discriminative performance. However, under annotator variability, naïve multitask optimization often fails not because the tasks are unrelated, but because their gradients compete within the shared representation. To make this competition explicit and controllable, we introduce \modelname{}, a representation-level adversarial gradient regularizer. Rather than performing parameter-level gradient surgery, \modelname{} uses each task's normalized adversarial direction in latent space as a geometric probe of task sensitivity and penalizes excessive angular alignment between task-specific adversarial directions. On a public TI-RADS dataset, our clinically guided multitask model with \modelname{} consistently improves risk stratification while maintaining segmentation quality compared to single-task training and conventional multitask baselines. Code and pretrained models will be released.
\end{abstract}

\section{Introduction}
Thyroid nodules are common, with ultrasound studies reporting detection in roughly 19--68\% of asymptomatic adults, while only 7--15\% are malignant~\cite{uludag2023thyroid}. Despite the low malignancy rate, accurate risk stratification is essential because it directly drives management decisions such as biopsy, surgery, minimally invasive ablation, or active surveillance~\cite{rodriguez2020risk,rago2022tirads}. Ultrasound (US) remains the primary modality for thyroid assessment due to its non-invasiveness, accessibility, low cost, and suitability~\cite{shi2022thyroid_ultrasound,jcu_thyroid_ultrasound_2023}. In clinical reporting, thyroid nodule assessment produces two coupled outputs: a nodule contour for measurement and documentation, and a TI-RADS category assigned from sonographic criteria (e.g., margins, echogenicity, shape, composition, echogenic foci) that guide biopsy versus follow-up~\cite{Tessler2017}. Both steps are reader-dependent. Different annotators may draw systematically different contours and assign different risk categories for the same case, meaning that the ``ground truth'' used for learning often reflects subjective interpretation rather than an unambiguous label. This annotator variability is not just noise; it creates inconsistent supervision that can destabilize training and harm generalization.

Deep learning has shown strong potential for thyroid US analysis, but common formulations only partially match this coupled workflow. Many works focus on classification alone, most often benign--malignant prediction using CNN-based models~\cite{liu2017thyroidcnn,zheng2023throid_unet,Buda2019}, while others focus on segmentation to enable downstream quantification~\cite{LI2023489,Zhou2025}. A widespread engineering choice is a sequential pipeline: segment first, then classify from the cropped or masked region. While intuitive, this design has two limitations: (i) segmentation errors and contouring style can propagate directly into risk prediction, and (ii) risk-related cues cannot shape the representation used for delineation, despite clinicians relying on overlapping evidence for both tasks. Joint learning is therefore natural, but shared-encoder multitask models often suffer from negative transfer when task gradients conflict, since naïvely optimizing multiple losses can lead to suboptimal shared representations ~\cite{NEURIPS2018_432aca3a,Yu2020PCGrad,liu2021conflictaverse}.

However, explicitly partitioning representations (e.g., Rep-MTL~\cite{RepMTL2025}) can be brittle when segmentation and TI-RADS grading rely on overlapping clinical cues under annotator variability, motivating a softer, geometry-based interference control within a shared representation.

In this work, we formulate thyroid US assessment as a coupled segmentation and TI-RADS risk stratification problem and learn both outputs jointly within a single, clinically aligned architecture. Beyond multitask learning, we make risk prediction clinically grounded by guiding part of the classification embedding toward a compact TI-RADS-aligned radiomics target during training, while retaining complementary deep features for discriminative power. Under annotator variability, multitask learning fails not because the tasks are unrelated, but because their gradients compete in the shared representation. To make this competition explicit and controllable, we introduce \modelname{}, a representation-level adversarial gradient regularizer inspired by adversarial sensitivity analysis~\cite{Goodfellow2015Explaining,Ilyas2019AdversarialFeatures}. Rather than performing parameter-level gradient surgery, \modelname{} uses each task's normalized adversarial direction in latent space as a geometric probe of task sensitivity and penalizes excessive angular alignment between task-specific adversarial directions. This encourages complementary (non-conflicting) task sensitivities while retaining clinically meaningful shared cues. We evaluate \modelname{} on a public TI-RADS dataset, where it improves risk stratification while maintaining segmentation compared to single-task and conventional multitask baselines.
\begin{figure}[t]
    \centering
    \includegraphics[width=\textwidth]{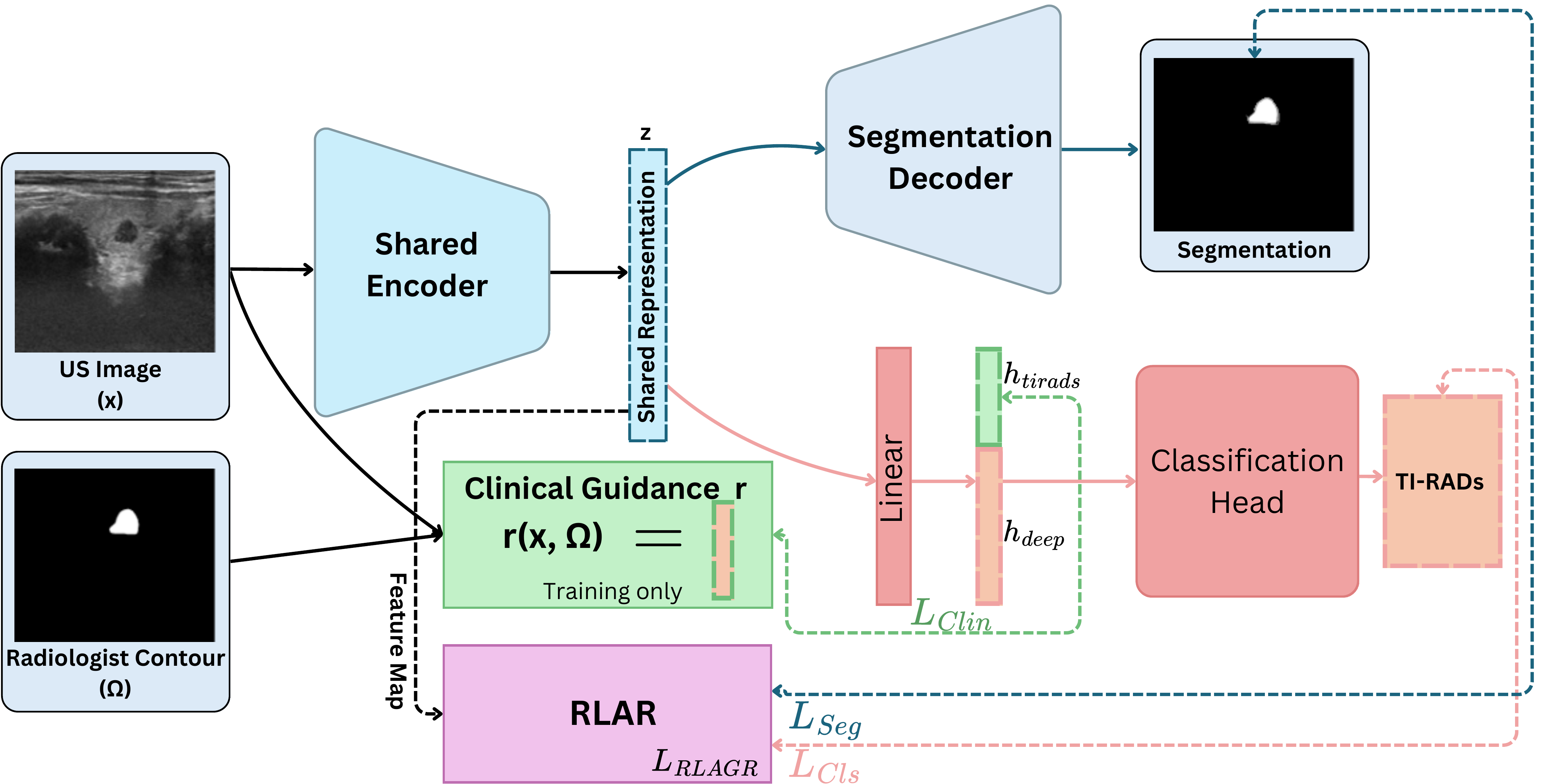}
    \caption{Overview of the proposed clinically guided multitask framework. A shared encoder jointly performs nodule segmentation and TI-RADS classification influenced by clinical guidance, while \modelname{} regularizes task interaction in the shared representation.}
    \label{fig:method}
\end{figure}
\section{Methodology}
As illustrated in Fig.~\ref{fig:method}, our framework uses a shared encoder to jointly support nodule segmentation and TI-RADS classification, while a training-only clinical guidance branch distills radiomics targets into an interpretable TI-RADS-aligned subspace.
The resulting shared representation is then decoded into a segmentation mask and mapped to TI-RADS logits through the classification head. Both segmentation and classification losses are used to calculate the gradients w.r.t shared representation and output Representation-Level Adversarial Regularization loss.
We model thyroid ultrasound assessment as the coupled prediction problem performed in routine reporting: nodule delineation for measurement and TI-RADS risk categorization. Given an ultrasound image batch
$x \in \mathbb{R}^{B \times 1 \times H \times W}$,
our network jointly outputs a segmentation map
$\hat{S} \in [0,1]^{B \times H \times W}$
and TI-RADS logits
$\hat{y} \in \mathbb{R}^{B \times C}$.
A shared encoder $E_\theta$ extracts a representation $z = E_\theta(x)$ that is consumed by two task heads:
\begin{equation}
\hat{S} = D_\theta(z), \qquad h = C_\theta(z) \in \mathbb{R}^{B \times K}, \qquad
\hat{y} = W h + b.
\end{equation}
This design enables the two tasks to share evidence, mirroring the clinical workflow where contour and TI-RADS cues are assessed jointly.

\noindent \textbf{2.1 Clinically Guided Multitask Architecture.}
A key goal is to make risk prediction clinically grounded rather than purely data-driven. To this end, we explicitly reserve part of the classification embedding for TI-RADS-aligned, interpretable descriptors:
\begin{equation}
h = \big[\,h_{\text{tirads}},\; h_{\text{deep}}\,\big],
\qquad
h_{\text{tirads}} \in \mathbb{R}^{B \times 13},\;
h_{\text{deep}} \in \mathbb{R}^{B \times (K-13)}.
\end{equation}
The subspace $h_{\text{tirads}}$ is trained to encode a compact set of radiomics-based surrogates for TI-RADS criteria, while $h_{\text{deep}}$ captures complementary non-linear features that improve discrimination. Both are used by the final classifier, allowing interpretability without sacrificing accuracy.

\noindent \textbf{2.2 TI-RADS-Inspired Clinical Guidance via Radiomics Distillation.}
We provide clinical guidance using a training-time only signal derived from radiomic measurements computed on the ground-truth nodule mask. For each training sample, we extract a 13-dimensional target vector
$r(x,\Omega) \in \mathbb{R}^{B \times 13}$
from the image $x$ and ground-truth segmentation
$\Omega \in \{0,1\}^{B \times 1 \times H \times W}$.
These targets are chosen to reflect TI-RADS-relevant cues and fall into three clinically motivated groups:

\noindent \textbf{(i) Shape and growth pattern:}
circularity, ellipticity, elongation, and aspect ratio (capturing regularity and the taller-than-wide criterion).

\noindent \textbf{(ii) Margin characteristics:}
edge sharpness and edge intensity contrast (quantifying boundary definition and lesion--parenchyma transition).

\noindent \textbf{(iii) Echogenicity and texture:}
first-order mean, entropy, kurtosis, and GLCM statistics (contrast, energy, correlation, IMC1), capturing brightness, heterogeneity, intensity outliers, and spatial gray-level dependencies.
We supervise the clinically guided subspace with an $\ell_2$ regression loss:
\begin{equation}
\mathcal{L}_{\text{clin}} =
\left\|h_{\text{tirads}} - r(x,\Omega)\right\|_2^2.
\end{equation}
Importantly, radiomics are never computed at inference time; they serve only as a distillation target that anchors the learned representation to clinically meaningful descriptors while keeping test-time deployment unchanged.

\noindent \textbf{2.3 Representation-Level Adversarial Regularization (\modelname{}).}
Jointly optimizing segmentation and classification introduces a practical failure mode: task gradients can compete in the shared representation, causing negative transfer. We address this by explicitly regularizing task interaction in latent space. Let $\tilde{z} \in \mathbb{R}^{B \times D \times H' \times W'}$ denote the shared bottleneck representation, and let the set of task losses be
$\mathcal{T} = \{\text{seg},\text{cls},\text{clin}\}$.
For each task $t \in \mathcal{T}$, we compute a normalized adversarial direction in representation space:
\begin{equation}
\delta_t = \epsilon \frac{\nabla_{\tilde{z}} \mathcal{L}_t}
{\|\nabla_{\tilde{z}} \mathcal{L}_t\|_2 + \varepsilon},
\end{equation}
which acts as a geometric probe of task sensitivity (the direction that increases $\mathcal{L}_t$ most). We flatten $\delta_t$ per sample to $f_t^{(i)} \in \mathbb{R}^{D H' W'}$ and quantify interference between task pairs via cosine similarity:
\begin{equation}
\mathrm{sim}_{t,t'}^{(i)} =
\frac{f_t^{(i)} \cdot f_{t'}^{(i)}}
{\|f_t^{(i)}\|_2 \, \|f_{t'}^{(i)}\|_2},
\quad t \neq t',\; i=1,\ldots,B.
\end{equation}

\noindent \modelname{} penalizes alignment using absolute pairwise similarities:
\begin{equation}
\mathcal{L}_{\text{\modelname{}}} =
\frac{\lambda_{\text{adv}}}{B}
\sum_{i=1}^{B}
\frac{2}{|\mathcal{T}|(|\mathcal{T}|-1)}
\sum_{t \neq t'}
\big|\mathrm{sim}_{t,t'}^{(i)}\big|, 
\end{equation}
where $\lambda_{adv}$ is the weight for the regularization term, 
this encourages complementary (non-conflicting) task sensitivities while preserving clinically meaningful shared cues, stabilizing multitask learning under inconsistent supervision.

\noindent \textbf{2.4 Training Objective.}
We optimize a weighted sum of segmentation Dice loss $\mathcal{L}_{\text{seg}}$,
weighted cross-entropy classification loss $\mathcal{L}_{\text{cls}}$,
the clinical guidance loss $\mathcal{L}_{\text{clin}}$, and \modelname{}:
\begin{equation}
\mathcal{L}_{\text{total}} =
\lambda_{\text{seg}} \mathcal{L}_{\text{seg}}
+ \lambda_{\text{cls}} \mathcal{L}_{\text{cls}}
+ \lambda_{\text{clin}} \mathcal{L}_{\text{clin}}
+ \mathcal{L}_{\text{\modelname{}}}.
\end{equation}

\section{Experiments and Results}
\noindent \textbf{Datasets}
We evaluate on two public thyroid ultrasound datasets: ThyroidXL~\cite{Duong2025ThyroidXL} and the Stanford AIMI Thyroid Ultrasound Cine-clip (AIMI)~\cite{thyroid_cine_clip}. ThyroidXL is an expert-labeled, pathology-validated benchmark (MICCAI 2025) with 11{,}545 B-mode images from 4{,}093 patients acquired at the Vietnam National Hospital of Endocrinology. We follow the official split and run 3-fold cross-validation within the training split (9{,}541 images / 3{,}275 patients); the TI-RADS distribution is TR1 (70), TR2 (942), TR3 (3{,}050), TR4 (2{,}949), and TR5 (2{,}530). AIMI contains cine-loop acquisitions from 167 patients with biopsy-confirmed nodules and TI-RADS scoring; we extract 15{,}093 labeled frames with a more imbalanced distribution (TR4 dominant): TR1 (52), TR2 (732), TR3 (4{,}084), TR4 (6{,}653), and TR5 (3{,}572). Unlike ThyroidXL’s static images, AIMI’s cine frames introduce temporal redundancy and higher intra-case variability. Unless stated otherwise, all models are trained on ThyroidXL; we report in-domain 3-fold CV on ThyroidXL and zero-shot external evaluation on AIMI (no fine-tuning).

\noindent \textbf{Implementation Details}
All experiments were run on a single 16\,GB GPU with 16 CPU cores using PyTorch 2.9.1 (CUDA 12.8). We optimize segmentation with Dice loss and TI-RADS classification with weighted cross-entropy, using AdamW (lr $1\times10^{-4}$, default $\beta$) and batch size 8. We perform 3-fold cross-validation on ThyroidXL and report mean $\pm$ standard deviation across folds. For multitask learning, we attach a lightweight classification head to a pretrained nnU-Net encoder at the bottleneck and train end-to-end. Training uses mixed precision for up to 500 epochs (250 training and 50 validation steps per epoch), selecting the best checkpoint by validation performance.

\noindent \textbf{3.3 Single-Task Baselines.}

\noindent \textbf{Classification.} 
We benchmark two strong single-task models, EfficientNet-B7~\cite{tan2020efficientnetrethinkingmodelscaling} and ConvNeXt~\cite{liu2022convnet2020s}. As shown in Table~\ref{tab:cls_all_merged}, performance drops markedly on AIMI, consistent with domain shift from static B-mode images (ThyroidXL) to cine-derived frames that exhibit temporal redundancy and higher intra-case variability. This gap suggests that TI-RADS prediction is less stable under cine acquisition and motivates methods that improve robustness to cross-domain appearance changes.
\begin{table}[t]
\centering
\caption{Nodule TI-RADS classification (mean $\pm$ std over 3 folds). Trained on ThyroidXL and evaluated on AIMI as external data.  \best{Bold} denotes the best value and \second{Underline} denotes the second-best value within each dataset/metric block.}
\label{tab:cls_all_merged}
\small
\setlength{\tabcolsep}{4pt}
\resizebox{\textwidth}{!}{%
\begin{tabular}{l c c c c c c}
\toprule
\multirow{2}{*}{\textbf{Method}} &
\multicolumn{3}{c}{\textbf{ThyroidXL}} &
\multicolumn{3}{c}{\textbf{AIMI (external)}} \\
\cmidrule(lr){2-4}\cmidrule(lr){5-7}
& \textbf{Precision} $\uparrow$ & \textbf{Recall} $\uparrow$ & \textbf{F1} $\uparrow$
& \textbf{Precision} $\uparrow$ & \textbf{Recall} $\uparrow$ & \textbf{F1} $\uparrow$ \\
\midrule
EfficientNet B7 (single-task)
& 0.575 $\pm$ 0.052 & 0.570 $\pm$ 0.041 & 0.561 $\pm$ 0.053
& 0.2728 $\pm$ 0.0166 & 0.2492 $\pm$ 0.0069 & 0.2492 $\pm$  0.0047 \\

ConvNeXt (single-task)
& 0.4822 $\pm$ 0.0123 & 0.4452 $\pm$ 0.0573 & 0.4136 $\pm$ 0.0388
    & 0.4139 $\pm$ 0.0560 & 0.3713 ± 0.0474 & \best{0.3853 ± 0.0362}
 \\

\midrule
Vanilla Multitask
& 0.613 $\pm$ 0.006 & 0.600 $\pm$ 0.017 & 0.600 $\pm$ 0.017
& 0.4400 $\pm$ 0.0529 & \second{0.3967} $\pm$ 0.0351 & 0.3633 $\pm$ 0.0764 \\
Clinically Guided
& \best{0.6467 $\pm$ 0.0058} &  \best{0.6200 $\pm$ 0.0100} & \best{0.6267 $\pm$ 0.0115}
& 0.4167 $\pm$ 0.0551 & 0.3633 $\pm$ 0.0473 & 0.3500 $\pm$ 0.0529 \\
+Rep-MTL
& 0.6233 $\pm$ 0.0153 & \second{0.6200 $\pm$ 0.0200} & \second{0.6167 $\pm$ 0.0252}
& \best{0.4700 $\pm$ 0.0361} & 0.3700 $\pm$ 0.0400 & \second{0.3700 $\pm$ 0.0300} \\
+\modelname{}
& \second{0.6300 $\pm$ 0.0000} &0.6100 $\pm$ 0.0200 & 0.6133 $\pm$ 0.0153
& \second{0.4167 $\pm$ 0.0321} & \best{0.4033 $\pm$ 0.0208} & \second{0.3700 $\pm$ 0.0458} \\
\bottomrule
\end{tabular}%
}
\end{table}

\begin{table}[t]
\centering
\caption{Segmentation performance (mean $\pm$ std over 3 folds). Trained on ThyroidXL and evaluated on AIMI as external data.  \best{Bold} denotes the best value and \second{Underline} denotes the second-best value within each dataset/metric block.}
\label{tab:seg_all_merged}
\small
\setlength{\tabcolsep}{4pt}
\resizebox{\textwidth}{!}{%
\begin{tabular}{l c c c c c c}
\toprule
\multirow{2}{*}{\textbf{Method}} &
\multicolumn{3}{c}{\textbf{ThyroidXL}} &
\multicolumn{3}{c}{\textbf{AIMI (external)}} \\
\cmidrule(lr){2-4}\cmidrule(lr){5-7}
& \textbf{Dice} $\uparrow$ & \textbf{IoU} $\uparrow$ & \textbf{HD95} $\downarrow$
& \textbf{Dice} $\uparrow$ & \textbf{IoU} $\uparrow$ & \textbf{HD95} $\downarrow$ \\
\midrule
Vanilla Multitask
&\best{0.856 $\pm$ 0.011} &\best{0.767 $\pm$ 0.013} & \best{41.05 $\pm$ 10.17}
& \best{0.6357 $\pm$ 0.0138} & \best{0.5073 $\pm$ 0.0143} & 206.56 $\pm$ 39.86 \\
Clinically Guided
& 0.8496 $\pm$ 0.0066 & 0.7595 $\pm$ 0.0078 & 47.10 $\pm$ 4.370
& 0.6220 $\pm$ 0.0098 & 0.4927 $\pm$ 0.0096 & \second{214.06 $\pm$ 9.620} \\
+Rep-MTL
& 0.8456 $\pm$ 0.0122 & 0.7548 $\pm$ 0.0133 & 51.09 $\pm$ 10.10
& 0.6180 $\pm$ 0.0108 & 0.4883 $\pm$ 0.0108 & 218.02 $\pm$ 13.78 \\
+\modelname{}
& \second{0.8556 $\pm$ 0.0018} & \second{0.7665 $\pm$ 0.0025} & \second{43.72 $\pm$ 4.150}
& \second{0.6277 $\pm$ 0.0177} & \second{0.4990 $\pm$ 0.0184} & \best{205.78 $\pm$ 17.24} \\
\bottomrule
\end{tabular}%
}
\end{table}

\noindent \textbf{3.4 Multitask.}
Building on nnU-Net’s strong single-task segmentation performance, we use a pretrained nnU-Net encoder as the shared multitask backbone and attach a lightweight TI-RADS classification head at the bottleneck. We train end-to-end with joint segmentation and classification losses, defining a \emph{Vanilla Multitask} baseline. Tables~\ref{tab:cls_all_merged} and~\ref{tab:seg_all_merged} report results on ThyroidXL and zero-shot external evaluation on AIMI.

\noindent \textbf{Classification.}
Table~\ref{tab:cls_all_merged} shows that, on ThyroidXL, the Clinically Guided multitask model achieves the best overall classification performance across precision, recall, and F1. \modelname{} attains the second-best precision with competitive F1 and exhibits near-zero fold-to-fold variance in precision, suggesting stable optimization under joint training.
On AIMI, all approaches degrade due to domain shift. Notably, \modelname{} achieves the highest recall (0.4033), improving sensitivity over all multitask variants while maintaining second-best precision and F1. This indicates that our representation-level adversarial regularization preferentially preserves recall under distributional shift, which is clinically important for reducing missed high-risk nodules.

\noindent \textbf{Segmentation.}
Table~\ref{tab:seg_all_merged} shows that the Vanilla Multitask baseline attains the best in-domain segmentation performance on ThyroidXL across Dice, IoU, and HD95. \modelname{} is consistently second-best on all three metrics, but with substantially lower fold-to-fold variance (e.g., Dice $\pm$0.0018 vs.\ $\pm$0.011), indicating more stable training under joint optimization.
On the external AIMI dataset, \modelname{} achieves the best HD95 (205.78) among multitask variants while maintaining second-best Dice and IoU. RLAR also reduces variance relative to Vanilla Multitask. Overall, the improved HD95 together with reduced variability suggests stronger boundary robustness and better cross-dataset generalization.

\noindent \textbf{3.5 Ablation Studies.}
We conduct ablation studies to analyze (i) where representation-level regularization is most effective within the shared encoder and (ii) how individual clinically guided features contribute to multitask performance and external generalization.

\noindent \textbf{Where to apply \modelname{}.}
In the main setting, \modelname{} forms adversarial task directions using gradients with respect to the shared bottleneck representation $\tilde{z}$ (Sec.~2.3). We ablate the representation level at which the regularizer is applied: (i) bottleneck (default), (ii) last encoder layer, (iii) mid encoder layer, and (iv) the mean direction over the last three encoder layers. Overall, applying \modelname{} at the \emph{bottleneck} yields the most consistent gains on ThyroidXL, while the \emph{mid encoder layer} performs best on AIMI (Tables~\ref{tab:ablation_layer_cls}--\ref{tab:ablation_layer_Seg}). In contrast, enforcing the regularizer only at earlier encoder features is less stable, likely because these features are more tightly coupled to low-level spatial details and are therefore more sensitive to adversarial perturbations.

\begin{table}[t]
\centering
\caption{Ablation on the representation layer used for \modelname{} (classification). Gradients are computed with respect to different encoder layers. Results reported as mean $\pm$ std over 3 folds. \best{Bold} denotes the best value and \second{Underline} denotes the second-best value within each dataset/metric block.}
\label{tab:ablation_layer_cls}
\small
\setlength{\tabcolsep}{4pt}
\resizebox{\textwidth}{!}{%
\begin{tabular}{l c c c c c c}
\toprule
\multirow{2}{*}{\textbf{Regularization Layer}} &
\multicolumn{3}{c}{\textbf{ThyroidXL}} &
\multicolumn{3}{c}{\textbf{AIMI (External)}} \\
\cmidrule(lr){2-4} \cmidrule(lr){5-7}
& \textbf{Precision} $\uparrow$ & \textbf{Recall} $\uparrow$ & \textbf{F1} $\uparrow$
& \textbf{Precision} $\uparrow$ & \textbf{Recall} $\uparrow$ & \textbf{F1} $\uparrow$ \\
\midrule
Last Encoder Layer
& \best{0.6333 $\pm$ 0.0058}
& 0.6033 $\pm$ 0.0058
& 0.6067 $\pm$ 0.0153
& \best{0.4567 $\pm$ 0.0351}
& 0.3567 $\pm$ 0.0586
& 0.3167 $\pm$ 0.0321 \\

Mid Encoder Layer
& \second{0.6300 $\pm$ 0.0100}
& 0.6033 $\pm$ 0.0058
& 0.6067 $\pm$ 0.0058
& \second{0.4300 $\pm$ 0.0436}
& \best{0.4200 $\pm$ 0.0400}
& \best{0.3933 $\pm$ 0.0493} \\

Mean (Last 3 Layers)
& 0.6267 $\pm$ 0.0115
& \best{0.6133 $\pm$ 0.0231}
& \best{0.6167 $\pm$ 0.0115}
& 0.4233 $\pm$ 0.0351
& 0.3767 $\pm$ 0.0153
& 0.3767 $\pm$ 0.0306 \\

Bottleneck (default)
& 0.6300 $\pm$ 0.0000
& \second{0.6100 $\pm$ 0.0200}
& \second{0.6133 $\pm$ 0.0153}
& 0.4167 $\pm$ 0.0321 & \second{0.4033 $\pm$ 0.0208} & 0.3700 $\pm$ 0.0458 \\

\bottomrule
\end{tabular}%
}
\end{table}
\begin{table}[t]
\centering
\caption{Ablation on the representation layer used for \modelname{} (segmentation). Gradients are computed with respect to different encoder layers. Results reported as mean $\pm$ std over 3 folds.  \best{Bold} denotes the best value and \second{Underline} denotes the second-best value within each dataset/metric block.}
\label{tab:ablation_layer_Seg}
\small
\setlength{\tabcolsep}{4pt}
\resizebox{\textwidth}{!}{%
\begin{tabular}{l c c c c c c}
\toprule
\multirow{2}{*}{\textbf{Regularization Layer}} &
\multicolumn{3}{c}{\textbf{ThyroidXL}} &
\multicolumn{3}{c}{\textbf{AIMI (External)}} \\
\cmidrule(lr){2-4} \cmidrule(lr){5-7}
& \textbf{Dice} $\uparrow$ & \textbf{IoU} $\uparrow$ & \textbf{HD95} $\downarrow$
& \textbf{Dice} $\uparrow$ & \textbf{IoU} $\uparrow$ & \textbf{HD95} $\downarrow$ \\
\midrule
Last Encoder Layer
& \second{0.8522 $\pm$ 0.0049}
& \second{0.7625 $\pm$ 0.0054}
& \second{45.74 $\pm$ 6.25}
& 0.6098 $\pm$ 0.0158
& 0.4802 $\pm$ 0.0148
& 235.48 $\pm$ 16.04 \\

Mid Encoder Layer
& 0.8379 $\pm$ 0.0064
& 0.7467 $\pm$ 0.0071
& 61.91 $\pm$ 7.87
& 0.6177 $\pm$ 0.0185
& 0.4884 $\pm$ 0.0164
& 223.10 $\pm$ 21.25 \\

Mean (Last 3 Layers)
& 0.8481 $\pm$ 0.0096
& 0.7578 $\pm$ 0.0120
& 51.11 $\pm$ 8.52
& \best{0.6296 $\pm$ 0.0089}
& \best{0.5028 $\pm$ 0.0077}
& \second{207.17 $\pm$ 18.83} \\

Bottleneck (default)
& \best{0.8556 $\pm$ 0.0018}
& \best{0.7665 $\pm$ 0.0025}
& \best{43.72 $\pm$ 4.15}
& \second{0.6277 $\pm$ 0.0177}
& \second{0.4990 $\pm$ 0.0184}
& \best{205.78 $\pm$ 17.24} \\

\bottomrule
\end{tabular}%
}
\end{table}

\noindent \textbf{Clinical-guidance feature importance.}
To quantify the contribution of each clinically guided feature, we perform a leave-one-feature-out analysis at inference time. For each feature channel $f_k$ in the guidance vector, we zero its corresponding classifier weight ($w_k \leftarrow 0$) while keeping all other weights fixed, and re-run inference. We then report the multitask classification performance difference with each feature disabled (Table~\ref{tab:ablation_feat_class}); larger drops indicate greater reliance on that feature under clinically guided multitask learning. On ThyroidXL, removing any single feature has only a minor effect, with small F1 decreases at most (largest drops: \textit{ellipticity} $-0.0030$, \textit{aspect\_ratio} $-0.0025$, and \textit{radiomics\_glcm\_correlation} $-0.0024$). This suggests that improvements in Table~\ref{tab:cls_all_merged} arise from complementary interactions among multiple features rather than dependence on a single cue. In contrast, AIMI is substantially more sensitive to feature removal. The largest degradations occur when disabling \textit{ellipticity} or \textit{radiomics\_glcm\_entropy} (both Recall $-0.0473$, F1 $-0.0265$), followed by \textit{radiomics\_kurtosis} (Recall $-0.0414$, F1 $-0.0213$). Removing \textit{radiomics\_entropy} or \textit{edge\_sharpness} yields comparable drops (approximately Recall $-0.040$, F1 $\approx -0.021$). Overall, these results indicate that external generalization on AIMI relies more heavily on texture- and shape-related guidance cues, whereas ThyroidXL remains comparatively robust to single-feature perturbations.

\begin{table}[t]
\centering
\small
\setlength{\tabcolsep}{4pt}
\caption{Clinical-guidance feature ablation (Classification). Each row reports the \emph{change} in performance relative to the full clinically guided model (all features enabled). Values are mean $\pm$ std over 3 folds; negative values indicate a performance drop.  \best{Bold} denotes the \emph{largest degradation} and \second{Underline} denotes the \emph{second-largest degradation} within each dataset/metric column.}
\label{tab:ablation_feat_class}
\resizebox{\textwidth}{!}{%
\begin{tabular}{l c c c c c c}
\toprule
& \multicolumn{3}{c}{\textbf{ThyroidXL}} & \multicolumn{3}{c}{\textbf{AIMI (External) }} \\
\cmidrule(lr){2-4} \cmidrule(lr){5-7}
\textbf{Disabled Feature}
& \textbf{Precision} $\uparrow$ & \textbf{Recall} $\uparrow$ & \textbf{F1} $\uparrow$
& \textbf{Precision} $\uparrow$ & \textbf{Recall} $\uparrow$ & \textbf{F1} $\uparrow$\\
\midrule
circularity
& \second{-0.0009 $\pm$ +0.0045} & -0.0002 $\pm$ -0.0044 & -0.0017 $\pm$ -0.0035
& +0.0032 $\pm$ +0.0216 & -0.0378 $\pm$ +0.0228 & -0.0148 $\pm$ +0.0047
\\
ellipticity
& -0.0004 $\pm$ +0.0042 & \best{-0.0018 $\pm$ -0.0042} & \best{-0.0030 $\pm$ -0.0024}
& +0.0008 $\pm$ +0.0236 &\best{ -0.0473 $\pm$ +0.0190} & \best{-0.0265 $\pm$ -0.0005}
\\
aspect\_ratio
& -0.0007 $\pm$ +0.0041 & \second{-0.0010 $\pm$ -0.0031} & \second{-0.0025 $\pm$ -0.0020}
& +0.0005 $\pm$ +0.0229 & -0.0398 $\pm$ +0.0231 & -0.0196 $\pm$ +0.0077
\\
edge\_sharpness
& +0.0002 $\pm$ +0.0042 & +0.0000 $\pm$ -0.0028 & -0.0016 $\pm$ -0.0018
& +0.0026 $\pm$ +0.0236 & -0.0398 $\pm$ +0.0251 & -0.0206 $\pm$ +0.0091
\\
edge\_intensity
& -0.0004 $\pm$ +0.0047 & -0.0003 $\pm$ -0.0032 & -0.0020 $\pm$ -0.0026
& +0.0025 $\pm$ +0.0220 &  -0.0395 $\pm$ +0.0252 &  -0.0191 $\pm$ +0.0083
\\
radiomics\_entropy
& +0.0003 $\pm$ +0.0043 & -0.0002 $\pm$ -0.0031 & -0.0017 $\pm$ -0.0020
& +0.0017 $\pm$ +0.0221 & -0.0404 $\pm$ +0.0243 & -0.0208 $\pm$ +0.0094
\\
radiomics\_mean
& +0.0000 $\pm$ +0.0047 & +0.0000 $\pm$ -0.0037 & -0.0016 $\pm$ -0.0027
& +0.0024 $\pm$ +0.0221 & -0.0396 $\pm$ +0.0246 & -0.0189 $\pm$ +0.0064
\\
radiomics\_kurtosis
& -0.0003 $\pm$ +0.0047 & -0.0008 $\pm$ -0.0032 & -0.0023 $\pm$ -0.0021
& \best{+0.0040 $\pm$ +0.0201} & \second{-0.0414 $\pm$ +0.0275} & \second{-0.0213 $\pm$ +0.0078}
\\
radiomics\_glcm\_contrast
& -0.0002 $\pm$ +0.0041 & +0.0000 $\pm$ -0.0032 & -0.0017 $\pm$ -0.0023
& +0.0020 $\pm$ +0.0215 & -0.0394 $\pm$ +0.0237 & -0.0185 $\pm$ +0.0062

\\
radiomics\_glcm\_energy
& +0.0009 $\pm$ +0.0046 & +0.0005 $\pm$ -0.0033 & \-0.0010 $\pm$ -0.0025
& \second{+0.0037 $\pm$ +0.0243} & -0.0395 $\pm$ +0.0247 & -0.0181 $\pm$ +0.0058

\\
radiomics\_glcm\_correlation
& \best{-0.0011 $\pm$ +0.0045} & -0.0006 $\pm$ -0.0028 & -0.0024 $\pm$ -0.0025
& +0.0005 $\pm$ +0.0229 & -0.0398 $\pm$ +0.0231 & -0.0196 $\pm$ +0.0077
\\
radiomics\_glcm\_entropy
& +0.0000 $\pm$ +0.0047 & +0.0000 $\pm$ -0.0037 & -0.0016 $\pm$ -0.0027
& +0.0008 $\pm$ +0.0236 & \best{-0.0473 $\pm$ +0.0190} &   \best{-0.0265 $\pm$ -0.0005}
\\
radiomics\_elongation2d
& -0.0007 $\pm$ +0.0044 & +0.0000 $\pm$ -0.0035 & -0.0021 $\pm$ -0.0023
& +0.0032 $\pm$ +0.0216 & -0.0378 $\pm$ +0.0228 & -0.0148 $\pm$ +0.0047
\\

\bottomrule
\end{tabular}
}
\end{table}

\section{Discussion and Conclusion}
We presented a clinically aligned multitask framework for thyroid ultrasound that jointly predicts nodule segmentation and TI-RADS categories using a shared encoder, mirroring clinical decision-making where morphology and boundary characteristics inform risk assessment. To ground classification in interpretable evidence, we distill a compact TI-RADS--aligned radiomics target into a dedicated embedding subspace during training, without adding any inference-time inputs or clinician burden. In-domain experiments on ThyroidXL show that multitask learning improves TI-RADS performance while maintaining strong segmentation accuracy, suggesting that contour supervision provides structural priors that support risk stratification. A key challenge in this setting is inconsistent supervision and annotator variability, which can make naïve parameter sharing unstable due to competing gradients in the shared representation. To address this, we introduced RLAR, a representation-level regularizer that explicitly moderates task interaction by discouraging excessive alignment between task-specific adversarial sensitivity directions. Unlike hard disentanglement or parameter-level gradient surgery, RLAR reshapes the shared latent space to reduce destructive interference while preserving beneficial sharing. Across folds, RLAR improves optimization stability and strengthens sensitivity under distribution shift, while maintaining segmentation quality. Our ablations further show that where interference is measured matters: bottleneck-level regularization is most reliable on ThyroidXL, while mid-level regularization is more effective on AIMI, likely reflecting the higher intra-case variability and temporal redundancy in cine-derived frames. Overall, these results indicate that lightweight, geometry-based interference control can stabilize clinically grounded multitask learning and improve robustness without changing the deployment workflow.

\bibliographystyle{splncs04}  
\bibliography{references}      

\end{document}